\documentclass[a4paper,10pt]{article}

\usepackage[utf8]{inputenc}
\usepackage[T1]{fontenc}
\usepackage[misc]{ifsym}

\usepackage{graphicx}
\usepackage{textcomp}
\usepackage{booktabs}
\usepackage{xcolor}
\usepackage[binary-units]{siunitx}
\usepackage{xspace}
\usepackage{url}
\usepackage{tabularx}
\usepackage{mathtools} %
\usepackage[noend]{algpseudocode}%
\usepackage[title]{appendix}
\usepackage{ifthen}

\usepackage{pgfplots}
\pgfplotsset{compat=1.18}

\definecolor{tolBblue}{HTML}{4477AA}
\definecolor{tolBcyan}{HTML}{66CCEE}
\definecolor{tolBgreen}{HTML}{228833}
\definecolor{tolByellow}{HTML}{CCBB44}
\definecolor{tolBred}{HTML}{EE6677}
\definecolor{tolBpurple}{HTML}{AA3377}
\definecolor{tolBgrey}{HTML}{BBBBBB}
\definecolor{tolHCwhite}{HTML}{FFFFFF}
\definecolor{tolHCyellow}{HTML}{DDAA33}
\definecolor{tolHCred}{HTML}{BB5566}
\definecolor{tolHCblue}{HTML}{004488}
\definecolor{tolHCblack}{HTML}{000000}

\usepackage[algpseudocodecmds,nocitepackage]{pauli_new}

\usepackage{redescriptions}

\newcommand{\vLHSHypiii}{\text{Hyp3}}

\newcommand{\vLHSHodiii}{\text{Hod3}}
\newcommand{\vLHSAL}{\text{AL}}
\newcommand{\vLHSOL}{\text{OL}}

\newcommand{\vRHSTIso}{\text{TIso}}
\newcommand{\vRHSPTotY}{\text{PTotY}}

\usepackage{custom_draw}

\renewcommand{\algname}[1]{\texttt{#1}}
\newcommand{\reremi}{\algname{ReReMi}\xspace}
\newcommand{\initialpairs}{\algname{InitialPairs}\xspace}

\newcommand{\reremibkt}{\algname{ReReMiBkt}\xspace}
\newcommand{\ownalgo}[1][]{%
  \algname{Fier%
    \ifthenelse{\equal{#1}{}}{}{\_#1}%
  }\xspace}
\newcommand{\ownalgoinit}{\ownalgo[init]}
\newcommand{\ownalgoext}{\ownalgo[ext]}
\newcommand{\ownalgofull}{\ownalgo[full]}

\renewcommand{\datasetname}[1]{\textsf{#1}}

\newcommand{\europeclimate}{\datasetname{EuroClim}\xspace}

\newcommand{\ethno}{\datasetname{Ethno}\xspace}
\newcommand{\cms}{\datasetname{CMS}\xspace}
\newcommand{\noisyeurope}{\datasetname{NoisyClim}\xspace}
\newcommand{\vaali}{\datasetname{VAA}\xspace}
\newcommand{\dentalworld}{\datasetname{DentalW}\xspace}
\newcommand{\dentalasia}{\datasetname{DentalA}\xspace}
\newcommand{\mamwc}{\datasetname{MammalsW}\xspace}

\newcommand\buk{\mathit{buk}}
\newcommand\nbuk{n_b}
\newcommand\minsupp{\supp_{\min}}

\newcommand\minout{\supp_{\mathrm{out}}}
\newcommand\sig{\mathit{sig}}

\newcommand\bands{b}
\newcommand\bandsjacc{b_J}
\newcommand\bandsham{b_H}
\newcommand\rows{r}
\newcommand\rowsjacc{r_J}
\newcommand\rowsham{r_H}
\DeclareGraphicsExtensions{.pdf,.png,.jpg}

\newcommand{\codeurl}{\url{https://github.com/maijuka/fier}}
\newcommand\doi[1]{\url{https://doi.org/#1}}

\usepackage{hyperref}
\hypersetup{colorlinks=true}

\begin{document}

\title{Fast Redescription Mining Using Locality-Sensitive Hashing}

\author{Maiju Karjalainen \and
Esther Galbrun  \and 
Pauli Miettinen}

\date{University of Eastern Finland \\ \texttt{firstname.lastname@uef.fi}}

\maketitle 
\begin{abstract}
Redescription mining is a data analysis technique that has found applications in diverse fields. The most used redescription mining approaches involve two phases: finding matching pairs among data attributes and extending the pairs. This process is relatively efficient when the number of attributes remains limited and when the attributes are Boolean, but becomes almost intractable when the data consist of many numerical attributes. In this paper, we present new algorithms that perform the matching and extension orders of magnitude faster than the existing approaches. Our algorithms are based on locality-sensitive hashing with a tailored approach to handle the discretisation of numerical attributes as used in redescription mining.

\end{abstract}

\section{Introduction}
\label{sec:introduction}

A redescription is a pattern that characterises roughly the same entities in two different ways, and redescription mining is the task of automatically extracting redescriptions from the input dataset, given user-defined constraints. Redescription mining has found applications in various fields of science, such as ecometrics. %
Ecometrics aims to identify and model the functional relationships between traits of organisms and their environments~\cite{Eronen10,Fortelius02}.
For instance, the teeth of large plant-eating mammals are adapted to the food that is available in their environment, which in turn depends on the climatic conditions, potentially allowing one to reason about the climate in the past based on the fossil record.

\begin{figure}[tbp]
  \includegraphics[height=4cm]{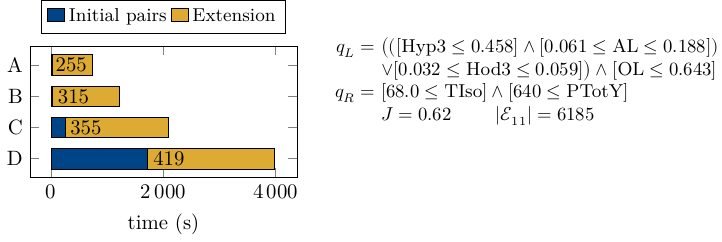}  
  \caption{Left: Running times on the \dentalworld dataset for finding initial pairs (blue) and extending pairs (yellow) using (A) the proposed algorithm (\ownalgofull), (B) \ownalgoinit for initial pairs and \reremi for extensions, (C) \reremi with pre-bucketing (\reremibkt) and (D) standard \reremi. The number within each bar indicates how many initial pairs were found. Right: Example redescription.}
  \label{fig:intro}
\end{figure}

To apply redescription mining in this context, the entities in the dataset represent localities, with two sets of attributes recording respectively the distribution of dental traits among species and the climatic conditions at each locality~\cite{galbrun17computational,liu23emergence}.
Galbrun et al.~\cite{galbrun17computational} mined redescriptions from this dataset using the \reremi~\cite{galbrun12black} algorithm in about $50$ minutes on a commodity laptop.
Replicating the experiment, we obtained a comparable time (bar D of Fig.~\ref{fig:intro}~(left)).

In contrast, the two top bars represent our proposed method, \ownalgo (Fast Initialisation and Extension of Redescriptions), based on locality-sensitive hashing (LSH). Our approach uses the same two-phase procedure as \reremi: it finds initial pairs in the first phase, and extends them in the second phase (see Sect.~\ref{sec:algorithm} for details). The top bar, \ownalgofull, uses a LSH-based approach for both phases, while the second bar, \ownalgoinit, uses LSH approach only for the first part. The third bar represents a straightforward speed-up of initial pair generation of \reremi that we use as a baseline. It is obtained by replacing on-the-fly discretisation of numerical attributes by discretisation as a pre-processing step.

Figure~\ref{fig:intro}~(left) shows that using \ownalgo the whole mining process is completed before the standard \reremi has even finished mining the initial pairs. The process is reduced from \qty{66}{\minute} to mere \qty{12}{\minute}; mining the initial pairs takes only \qty{25}{\second}.

\ownalgo, being significantly faster than the existing methods, allows the use of redescription mining on even larger datasets. Alternatively, the speedup affords more responsive interactive redescription mining~\cite{galbrun18mining} and quickly testing parameter and constraint combinations to `get a feel' for what kind of setup to use with the more exhaustive \reremi algorithm. The price \ownalgo has to pay for its speed is its probabilistic nature: unlike \reremi, it does not guarantee to return the best initial pairs nor the (locally) best extensions.

\paragraph{Redescription mining.}
The input of redescription mining is a pair of \emph{data tables} which we refer to as the left-hand side and right-hand side, denoted respectively $\Table_L$ and $\Table_R$, over the same \emph{entities}, denoted $\Entities$.
Then, a redescription is a pair of \emph{queries}, denoted $\lquery$ and $\rquery$, consisting of \emph{literals} over the attributes of the corresponding table, combined with logical conjunction and disjunction operators, possibly involving negations.
The set of entities that satisfy both queries, only the left-hand side query, only the right-hand side query, and neither of them are denoted respectively $\Exx$, $\Exo$, $\Eox$ and $\Eoo$. The set $\Exx$ is also called the support of the redescription, and more in general, we call \emph{support} ($\supp$) of a literal or query the set of entities that satisfy it.
The Jaccard index between the sets of entities that satisfy either query is called the \emph{accuracy} of the redescription (we use the terms Jaccard and accuracy interchangeably) and is defined
\[\jacc(\lquery, \rquery) = \frac{\Nxx}{\Nxo + \Nxx + \Nox}\,.\]
Since it is meant to provide two ways to characterise roughly the same entities, accuracy is the main measure of the quality of a redescription.
Further constraints that can be applied to redescriptions include a maximum threshold on the number of attributes they involve, since long queries might be difficult to interpret, and minimum thresholds on the number of entities that satisfy both queries and that satisfy neither, since redescriptions that characterise too few or too many of the entities are typically not considered interesting.

Going back to the application in ecometrics, the right-hand side query of the example redescription shown in Fig.~\ref{fig:intro}~(right) characterises the climatic conditions encountered at the locality, in this case requiring high isothermality ($\vRHSTIso{}$) and high annual precipitation ($\vRHSPTotY{}$).
The left-hand side query ($\lquery$) is more complex and involves four literals over dental traits, requiring a limited prevalence of hypsodont species ($\vLHSHypiii{}$) and fairly low fraction of species with acute lophs ($\vLHSAL{}$), or a very low fraction of hypsohorizodont species ($\vLHSHodiii{}$), but allowing up to a rather large fraction of species with obtuse lophs ($\vLHSOL{}$).
Hypsodont and hypsohorizodont species refer to species with elongated teeth respectively along the vertical and the horizontal dimension, while acute and obtuse lophs refer respectively to the presence of sharp and blunt edges on the tooth across the chewing direction. 
In the considered dataset, $\Nxx=6\,185$ of the \num{28886} localities satisfy both queries, representing about $\jacc=\qty{62}{\percent}$ of the \num{10011} localities that satisfy at least one of the queries.

\paragraph{Related work.}
\label{sec:related-work}

In the two decades since redescription mining was introduced~\cite{ramakrishnan04turning}, various algorithms have been proposed for this task, including based on decision tree induction~\cite{ramakrishnan04turning,zinchenko15mining,mihelcic17framework}, itemset mining~\cite{gallo08finding}, as well as iterative greedy heuristics~\cite{gallo08finding,galbrun12black}. The \reremi algorithm~\cite{galbrun12black} belongs to the latter family.

Other lines of work have focused for instance on selecting a good set of redescriptions~\cite{kalofolias16from}, designing an interactive  tool~\cite{galbrun18mining} and providing differentially private methods for redescription mining~\cite{MihelcicM23,KarjalainenGM22}.
Meanwhile, redescription mining has been used in applications from domains as diverse as medicine~\cite{Mihelcic17using}, political sciences~\cite{galbrun16analysing} and palaeontology~\cite{galbrun17computational,liu23emergence}.

Also relevant here is the work of Meeng and Knobbe~\cite{meeng21real} studying the impact of different discretisation strategies on subgroup discovery, a task very similar to redescription mining.

Locality-sensitive hashing was introduced for efficient nearest neighbour search~\cite{indyk98approximate}. It has proven to be very useful for various tasks, such as collaborative filtering~\cite{das07google}, clustering~\cite{cochez15twister} and privacy preservation~\cite{fernandes21locality}. The work most related to the present work is the early application to faster association rule mining~\cite{cohen2001}. 

\section{The Algorithm}
\label{sec:algorithm}

We divide our full algorithm into two parts for the two phases: \ownalgoinit computes the initial pairs and \ownalgoext the extensions. The full algorithm is called \ownalgofull. 
Before presenting the algorithms, we provide short primers on the \reremi algorithm and on locality-sensitive hashing.

\subsection{The \reremi algorithm}
\label{sec:reremi-algorithm}

The algorithm on which we build, \reremi~\cite{galbrun12black}, mines redescriptions in two main phases. In the first phase, the \initialpairs method returns candidate redescriptions with a single literal on either side. In the second phase, up to a chosen number of the best initial pairs are considered. Each is extended in turn through a greedy process that iteratively appends a literal to either of the queries to produce more refined and accurate redescriptions.
The \reremi algorithm can handle Boolean, categorical and numerical attributes. In particular, when forming literals for a numerical attribute, it performs on-the-fly discretisation, trying to find the interval that yields the best accuracy for the considered candidate, rather than using pre-determined buckets. 

Going a bit further into technical details, %
the way \reremi finds the initial pairs is simple but time consuming: Iterate over all pairs of attributes, one from either side; For each such pair, consider the possible pairs of literals over the two attributes; Compute the support and accuracy of the corresponding redescription and keep the best matches that satisfy support constraints.

Given a redescription to extend, \reremi considers every available literal and the candidate extensions obtained by appending it (or its negation if allowed) to the current query on the corresponding side, with either the disjunction or the conjunction operator. The support and accuracy of each candidate extension are computed, the best one-step extensions are selected and extended in turn, for each one considering again all the possible ways to extend it with the remaining available literals. Starting with an initial pair, this greedy extension process is applied until no further improvement (in terms of accuracy) can be found or the maximum length of the queries (in terms of the number of involved literals) has been reached. The algorithm then proceeds with the next initial pair.

Computing the support and accuracy for a pair of literals involves computing the cardinality of the intersection and union of the sets of entities satisfying either literal.
Similarly for computing the support and accuracy for an extension, although it has been shown~\cite{galbrun12black} that not all entities can change status in the support of an extension. For instance, when extending a redescription by adding a literal to the left-hand side with a conjunction, entities that do not satisfy the current left-hand side query will not satisfy the extended one either. This fact is used to quickly identify the entities that need to be considered when performing the set operations to determine the quality of an extension.

\subsection{Primer on LSH}
\label{sec:primer-lsh}

The idea of Locality-Sensitive Hashing (LSH) is to calculate hash values for input items so that similar items get the same hash value with a high probability. The type of hash function depends on the similarity measure used. In this work we use two measures of similarity between binary vectors representing support sets, namely the Jaccard similarity and the Hamming distance. %

When using the Jaccard similarity, the corresponding hashing technique is called minhashing~\cite{broder1998min}. To compute minhash values, we use $k$ random hash functions ${h_1,\ldots,h_k}$ that map the row indices to values between $0$ and the maximum number of rows. To calculate the minhash signature of a vector, we consider the indices of all the rows containing a $1$. For every hash function we obtain the hash values for all of these rows and take the smallest value. These concatenated smallest hash values make up the signature of the vector~\cite{leskovec2020mining}.

When using the Hamming distance, we calculate the length-$k$ signature of the vector as the values of the binary vector in $k$ random indices.

To match two binary vectors based on their signatures, the signatures are divided into $\bands$ smaller sections called bands, each containing $\rows$ hash values (total $k=\rows \cdot \bands$). Two vectors are paired if all $\rows$ hash values match in at least one band.

The parameters $\bands$ and $\rows$ determine the matching probability, i.e. the probability with which two vectors will be paired. If the similarity of the vectors is $p$, then the matching probability is $1-(1-p^r)^b$. %
The relation between the similarity and the matching probability follows an S-curve; its threshold point can be approximated as $p_T=(1/\bands)^{1/\rows}$~\cite{leskovec2020mining}. With this approximation, only two parameters need to be set and the third one can be calculated accordingly. To increase the chances of finding pairs with an accuracy above a desired threshold, one can use a slightly lower value, whereas a slightly higher value yields a faster algorithm finding fewer pairs.

\subsection{Finding Initial Pairs}
\label{sec:initial-pairs}

For simplicity we first focus on Boolean attributes.
The pseudocode for the algorithm is given in Algorithm \ref{alg:initial_pairs}. 
Starting with the left-hand side dataset $\Table_L$, we go through the literals and obtain $\rowsjacc$ minhash signatures for each of the corresponding binary support vectors. Literals with the same signature get hashed into the same bin, and we keep track of which side each literal belongs to.

We go through the right-hand side dataset $\Table_R$ similarly, except that we discard signatures that point to an empty bin.
After calculating signatures for all literals, we go through each signature bin and form candidate pairs between the literals from $\Table_L$ and $\Table_R$ hashed into that bin.

We repeat this process $\bandsjacc$ times, calculating a different minhash signature for the literals for each band. If the same two literals already formed a pair in a previous band, we do not consider it again.
After the candidate pairs have been formed, we calculate their actual accuracy and support, and we filter out any candidate that does not satisfy user-defined support and accuracy thresholds.

\begin{algorithm}[tb]
  \caption{Generating initial pairs from Boolean attributes}\label{alg:initial_pairs}
  \begin{algorithmic}[1]
    \small
    \Input Data $\Data = (\Table_L, \Table_R)$, number of bands $\bandsjacc$, width of bands $\rowsjacc$
    \Output A set of initial redescriptions $\col{R} = \{(\lquery, \rquery)\}$
    \Function{\ownalgoinit}{\Data, b, r} 
    \For{$band=1,\ldots, b$} 
      \State $B \gets \emptyset$
      \State generate $r$ hash functions $h_1, \ldots, h_r$ that permute the row indexes
      \For{side in $\Data$}
      \For{attribute $a$ in side}
      \State $\sig \gets \bigl(\min\bigl(h_1(\supp(a))\bigr),\ldots,\min\bigl(h_r(\supp(a))\bigr)\bigr)$
      \State $B[\sig]\gets B[\sig] \cup \{a\}$
      \EndFor
      \ForAll{$\sig \in B$}
      \State $\col{R} \gets \col{R}\cup \{$pairs of literals $(a_L, a_R)$ in $B[\sig]$, $a_L\in\Table_L,a_R\in\Table_R \}$ 
      \EndFor
      \EndFor
     \EndFor
  \State \textbf{return} $\col{R}$
  \EndFunction
  \end{algorithmic}
\end{algorithm}

Building initial pairs from categorical attributes is slightly more complicated, as we can create several different literals by considering the different categories, separately or combined.
We first consider the literals obtained by considering each category of an attribute $a$ separately %
and calculate their minhash signatures the same way as for Boolean literals. Next, we create combinations of the categories by joining them with the disjunction operator, e.g. $[a=c_1\lor a=c_2]$. The number of categories that are combined can be limited with a user-defined parameter. Since we already calculated the signatures for the separate categories, we can obtain the signature for a combination by simply taking the smallest minhash value among the categories included in the combination.

Creating literals out of numerical attributes is similar to categorical attributes, but by specifying intervals such as $[a\leq x]$ or $[x\leq a\leq y]$, instead of categories. We determine the intervals in two steps. First, we discretise the attribute values into $\nbuk$ small buckets $[\buk_l \leq y \leq \buk_u]$ (line \ref{alg:num:buckets} in Algorithm \ref{alg:initial_pairs_num}) and calculate signatures for the literals corresponding to the separate buckets. Second, we create extended intervals by combining consecutive buckets, and calculate the signature for an extended interval by taking the smallest minhash values of the buckets included in the interval (line \ref{alg:num:sig_buckets}). We use support thresholds $\minsupp$ and $\minout$ to limit the size of intervals so that they do not cover too few or too many entities. The $\minout$ threshold defines the minimum number of entities not covered by the literal.

The discretisation of attribute values into buckets can be done in different ways.  We found that the bucketing method did not have a significant impact on the running time nor on the quality of the results. In our experiments we used equal-height binning, where each bucket covers a similar number of entities.

To avoid creating multiple pairs that are very similar to each other, we filter out any subinterval that has the same signature as the interval it is contained in. We do this by keeping track of the signatures we have seen for an attribute (within one band) as we iterate over the intervals.

We iterate over the intervals by progressively narrowing them down.
Considering the ordered bucket edges $\buk_0, \buk_1, \ldots, \buk_{\nbuk}$, we start by setting $\buk_0$ as the lower and $\buk_{\nbuk}$ as the upper bound of the interval. We lower the upper bound by removing buckets from the top $\buk_{\nbuk-1},\buk_{\nbuk-2},\ldots$ until the support criteria is met. We calculate the signature as the smallest minhash values of the buckets contained in the interval. We lower the upper bound until the support of the interval is below the minimum support criterion. Then, we set $\buk_1$ as the lower bound and repeat the process until we have gone through all lower bounds. We filter out subintervals of already seen intervals during this process, by discarding those intervals whose signature we have already seen, and whose upper bound is smaller than the largest seen upper bound for that signature (line \ref{alg:num:filter}).

\begin{algorithm}[tb]
  \caption{Generating initial pairs from numerical attributes}\label{alg:initial_pairs_num}
  \begin{algorithmic}[1]
    \small
    \Input Data $\Data = (\Table_L, \Table_R)$, number of bands $\bandsjacc$, width of bands $\rowsjacc$, number of buckets $\nbuk$, min. nb. of entities not in the support $\minout$, minimum support $\minsupp$
    \Output A set of initial redescriptions $\col{R} = \{(\lquery, \rquery)\}$
    \Function{\ownalgoinit}{\Data, b, r, \nbuk, \minout, \minsupp}
    \For{$band=1,\ldots, b$}  
    \For{data table in $\Data$}
    \For{attribute $a$ in data table}
    \State $[\buk_0, \buk_2, \ldots, \buk_{\nbuk}] \gets \Call{Discretise}{a, \nbuk}$ \label{alg:num:buckets}
    \For{$k=1,\ldots,\nbuk$}
    	\State Signatures$[k,:]\gets \rowsjacc$ minhash values for $[\buk_{k-1} \leq a \leq \buk_{k}]$
    \EndFor
    \State $S\gets\emptyset$
    \For{$l=0,\ldots,\nbuk-1$} \label{alg:num:filter_loop}
       \State $u \gets \nbuk$
       \State \textbf{while} $\supp([\buk_l\leq a\leq\buk_u])>\abs{\Entities} - \minout$ \textbf{do} $u \gets u-1$
    	\While{$\supp([\buk_l\leq a\leq\buk_u])\geq\minsupp$}
    		\State $\sig\gets\min_{i=l+1}^u($Signatures$[i,\colon])$ \label{alg:num:sig_buckets}
  			\If{$u>S[\sig]$} \label{alg:num:filter}
  				\State $B[\sig]\gets B[sig] \cup \{[\buk_l \leq a \leq \buk_u]\}$
  				\State $S[\sig]\gets u$
  			\EndIf
  			\State $u \gets u-1 $
    	\EndWhile
    \EndFor
    \EndFor
    \EndFor
     \ForAll{$\sig \in B$}
      \State $\col{R} \gets \col{R} \cup \{$pairs of literals $([\buk_l \leq a_L \leq \buk_u], [\buk'_l \leq a_R \leq \buk'_u])$ in $B[sig]$, $a_L\in\Table_L,a_R\in\Table_R\}$ 
      \EndFor
    \EndFor
  \State \textbf{return} $\col{R}$
  \EndFunction
  \end{algorithmic}
\end{algorithm}

\subsection{Extending Initial Pairs}
\label{sec:extension}

\paragraph{Computing signatures for literals.}
We start the process by computing the Hamming signatures for all data columns. For this we first randomly select indices over the number of rows in the data, so that we have $\bandsham$ sets of length $\rowsham$ random indices to create all signatures from. Note that these are different parameters than $\bandsjacc$ and $\rowsjacc$ used with the initial pair mining. For each literal to extend with, we store $\bandsham$ sets of length $\rowsham$ signatures.

The discretisation of the numerical attributes is done differently from the initial pair search. We start with a small number of buckets to discretise the values into and compute the Hamming signature corresponding to each bucket. Next, we double the number of buckets, perform the discretisation again and compute new signatures. We repeat this process a fixed number of times. We store the signatures for each bucket of each attribute, but we do not store the actual intervals, as they will be re-calculated later.  This approach has only a modest effect on the running time, as we only need to compute the signatures for literals once. On the other hand, it sidesteps the issue of choosing the correct bucketing by using many different ones.

\paragraph{The target vector.}
As mentioned earlier, only some entities in the data matter when extending a redescription. Let us consider a redescription $(\lquery, \rquery)$ which we want to extend on the right side with a conjunction. To improve $\jacc(\lquery, \rquery)$, we want to find a literal that is satisfied by entities in $\Exx$ (so that the numerator does not shrink) and not by entities in $\Eox$ (so that the denominator shrinks). Entities in $\Exo$ and $\Eoo$ do not matter, and we call these rows the `don't-care' rows. In terms of LSH, we can see this as having a target vector that has $1$s in rows corresponding to $\Exx$ and $0$s in rows corresponding to $\Eox$. We can easily find data columns that match this target vector using the Hamming distance restricted to these rows (we use Hamming instead of Jaccard to reward matches on $1$s and on $0$s equally). Similarly, when extending the left side with a disjunction, we want to find literals that have $1$s in rows corresponding to $\Eox$ and $0$s in rows corresponding to $\Eoo$, with $\Exx$ and $\Exo$ being the `don't-care' rows. When extending the right side either by conjunction or disjunction the target vectors are otherwise same as with the left side, except that $\Exo$ and $\Eox$ switch places. As these subsets of rows are different for each redescription to be extended, we would have to re-compute the signatures for all data columns considering every candidate redescription and both connectives used to extend it on each side. This would clearly cancel the speed benefit. Instead, we replace the `don't-care' rows in the target vector with $0$s when extending with conjunction and with $1$ when extending with disjunction. This way we only need to compute the signatures for the data columns once and we can use the same signatures for all extensions.

\paragraph{Extending redescriptions.}
The algorithm for extending the redescriptions, denoted \ownalgoext, is presented in Algorithm~\ref{alg:extend}. We start by storing the initial pairs in a priority queue $Q$, using accuracy as the key. We always expand the redescription with the current highest accuracy. A redescription can be extended on either of its sides, with a conjunction or a disjunction operator. The corresponding target vectors are computed by \textsc{ComputeTargetVector}. \textsc{ComputeCandidates} uses LSH to find the columns that match the target vectors, recording also the associated side and operator. The same column can match multiple times (in different bands, or different buckets for a numerical attribute), only the first match is stored. The actual extensions are then computed using the same approach as with \reremi (i.e.\ using the cut-point method~\cite{galbrun12black} for numerical attributes) and the best extension is stored and pushed back to $Q$ if it has not already been extended to maximum length. Compared to \reremi, the speed of \ownalgoext benefits from not trying every column as a candidate extension, but only those that match with LSH. 

\begin{algorithm}[tb]
  \caption{\ownalgoext: Extend redescriptions}\label{alg:extend}
  \begin{algorithmic}[1]
    \small
    \Input Initial pairs $\col{P}$, buckets with signatures for each possible extension $B$, maximum length of a redescription $t$
    \Output A set of extended redescriptions $\col{R}$
    \Function{\ownalgoext}{P,B,t} 
    \State $Q\gets$ heapify the list of initial pairs with accuracy as the key; $R\gets\emptyset$
    \While{$Q\neq\emptyset$}
      \State $(\lquery, \rquery) \gets Q.\mathrm{pop()}$
        \State $V\gets$ \Call{ComputeTargetVectors}{$(\lquery,\rquery)$}
        \State $C\gets$ \Call{ComputeCandidates}{$V, B$}
        \For{(column, side, operator) $\in C$}
          \State $E \gets$ \Call{ComputeExtensions}{$(\lquery,\rquery)$, column, side, operator}
        \EndFor
        \If{$E\neq\emptyset$}
          \State $\mathrm{best} \gets$ extension with the highest Jaccard
          \State $\col{R} \gets \col{R} \cup \{\mathrm{best}\}$
          \State \textbf{if} $\operatorname{len}(\mathrm{best}) < t$ \textbf{then}
            $Q.\mathrm{push}(\mathrm{best})$
        \EndIf
    \EndWhile
  \State \textbf{return} $\col{R}$
  \EndFunction
  \end{algorithmic}
\end{algorithm}

\subsection{Time Complexity}
\label{sec:time-complexity}

The time complexity for mining initial pairs from Boolean attributes is 
$O(N\cdot{}b\cdot{}r\cdot{}\abs{\Entities} + n_L\cdot{}n_R)$, where $\abs{\Entities}$ is the number of entities, $N$ is the total number of attributes, $b\cdot{}r$ is the total number of hash functions in LSH, $n_L$ is the largest number of left-hand side literals hashed to the same bin, and $n_R$ is the largest number of right-hand side literals hashed to the same bin. The first term is for computing the signatures. The second term is for building the pairs; in the worst case it is quadratic, but in practise much smaller. 

The time complexity for numerical attributes is 
$O\bigl(N(\abs{\Entities}\log \abs{\Entities}+\nbuk\cdot{}b\cdot{}r\cdot{}\abs{\Entities}+\nbuk^2) + n_L\cdot{}n_R\bigr)$; assuming $\nbuk$, $b$, and $r$ are constants, this becomes $O(N\abs{\Entities}\log\abs{\Entities}+n_L\cdot{}n_R)$. Bucketing with equal-height binning takes $O\bigl(N(\abs{\Entities}\log \abs{\Entities})\bigr)$. We can think of each bucket as creating a new literal, so creating the signatures takes $O(N\cdot{}\nbuk\cdot{}b\cdot{}r\abs{\Entities})$. Subinterval filtering (Algorithm \ref{alg:initial_pairs_num} line \ref{alg:num:filter_loop}) takes $O(\nbuk^2)$ for each of the  $N$ attributes. Similarly to Boolean attributes we assume at most $n_L$ and $n_R$ literals hashed to the same bin. The time complexity for the categorical attributes falls between the Boolean and numerical attributes.

The time complexity for mining the extensions is $\tau\cdot n_c\cdot T_R$, where $\tau$ is the number of extensions done (at most the number of initial pairs times the maximum length of a redescription), $n_c$ is the maximum number of candidate extensions for a target vector (at most the number of literals in the data but typically much fewer), and $T_R$ is the time \reremi takes to compute an extension for a given redescription and a literal.

\section{Experimental Evaluation}
\label{sec:experiments}

The experimental evaluation asserts that the results obtained by \ownalgo are comparable to \reremi in quality and that \ownalgo is significantly faster than \reremi. It is split into three parts: generating the initial pairs (Sect.~\ref{sec:exp-initial}), extending initial pairs (Sect.~\ref{sec:extend-init-pairs}) and the full system (Sect.~\ref{sec:exp-full}). We also test the sensitivity of \ownalgo to the parameters of locality-sensitive hashing. We did not test the effects of standard redescription mining parameters, such as $\minsupp$, that are common to all algorithms of this family and not a property of our proposed approach. %

\subsection{Experimental Setup}
\label{sec:experimental-setup}

We used twelve different datasets. Their properties are listed in Table~\ref{tab:datasets}.

\begin{table}[tbp]
  \centering
  \caption{dataset properties.}\label{tab:datasets}
\begin{tabular}{
  @{}
  l @{\hspace{6ex}}
  R@{\hspace{4ex}}
  R @{\hspace{1ex}}
  l @{\hspace{3ex}}
  R @{\hspace{1ex}}
  l@{}
}
  \toprule
  dataset & {\abs{\Entities}, \text{entities}}  & %
       \multicolumn{2}{l}{$\Table_{\iLHS}$ attributes} & \multicolumn{2}{l}{$\Table_{\iRHS}$ attributes} \\
  \midrule
  \europeclimate & 2\,575 & 12 & numerical & 12 & numerical \\
  \noisyeurope & 2\,575 & 36 & numerical & 36 & numerical \\
  \mamwc & 54\,013 & 48 & numerical & 4\,754 & Boolean\\
  \ethno & 1\,267 & 23 & numerical and categorical & 90 & numerical \\
  \dentalworld & 28\,886 & 11 & numerical & 19 & numerical \\
  \dentalasia & 6\,404 & 7 & numerical & 19 & numerical \\
  \vaali & 1\,656 & 9 & categorical & 107 & categorical \\
  \cms{$_d$} & d & 152 & numerical & 452 & numerical \\
  \bottomrule
\end{tabular}
\end{table}

The \mamwc dataset contains information about which mammal species inhabit which areas of the world on one side, and climate information on the other side~\cite{hijmans05very}.
That same climate data restricted to Europe, with monthly average temperatures on one side and monthly total precipitations on the other, constitutes the \europeclimate dataset. \noisyeurope contains three copies of the columns of \europeclimate. The first copy is as is, the second and third copies are perturbed by adding noise distributed following $\mathcal{N}(0, 1)$ and $\mathcal{N}(0, 3)$ respectively.

\dentalworld is the dataset from the ecometrics study~\cite{galbrun17computational} presented in the introductory example. One side contains dental traits aggregated over the resident large herbivorous species whereas the other side contains climate information, at localities across the world.
\dentalasia is a similar dataset, restricted to Southeast Asia and China and considering a slightly different set of dental traits~\cite{liu23emergence}.

\ethno is based on Murdock's Ethnographic Atlas~\cite{murdock67ethnographic}. It contains ecological attributes on the left-hand side, and nominal attributes recording cultural features of various tribes on the right hand side. 
\vaali~\cite{hs11vaalikone} contains the background information and answers to questions regarding political opinions of candidates to the Finnish parliamentary elections of 2011, as collected by an online voting advice application (VAA).

Finally, the \cms data are based on the synthetic data released by the Centers for Medicare and Medicaid Services.\!\footnote{\url{https://www.cms.gov/Research-Statistics-Data-and-Systems/Downloadable-Public-Use-Files/SynPUFs/DE_Syn_PUF}} We used five versions of the data, all having the same attributes, but increasing number of rows (\num{7199}, \num{14414}, \num{29136}, \num{58204} and \num{116395}). %
The differently sized versions of \cms are indicated with the numbers of rows in subscript where relevant. 

All algorithms are implemented in Python and the experiments (except when mentioned otherwise) are run with Python 3.6.8 on a machine with $2$ AMD EPYC 7702 processors with $64$ cores each and \SI{1}{\tera\byte} of main memory. All experiments were run single-threaded. For \reremi, we used the publicly available version.\!\footnote{\url{https://pypi.org/project/python-clired/}} Our code is publicly available.\!\footnote{\codeurl}

Unless otherwise noted, all comparative experiments were run with parameters $\bandsjacc=40$, $\rowsjacc=10$, $\minsupp=0.1\abs{\Entities}$, $\minout=0.3\abs{\Entities}$ and with $\nbuk=40$ buckets for numerical attributes. An exception to this are the \dentalworld and \dentalasia datasets, for which we used the parameters $\bandsjacc=40$, $\rowsjacc=5$, $\nbuk=50$ and $\minsupp=0.01\abs{\Entities}$. For \dentalworld we set $\minout$ to $0.6\abs{\Entities}$ and for \dentalasia to $0.3\abs{\Entities}$. When using \ownalgo, $\minout$ was further multiplied by $1.15$ for \dentalworld and $1.2$ for \dentalasia.
Further experimental results are presented in Appendix~\ref{sec:app-exper-eval}.

\subsection{Finding Initial Pairs}
\label{sec:exp-initial}

\begin{figure}[tbp]
  \centering
  \includegraphics{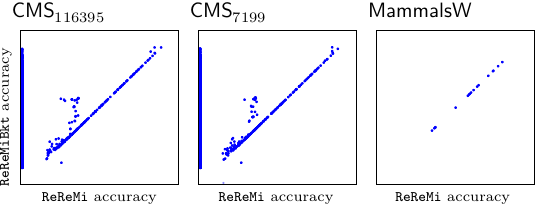}
  \caption{Comparing the accuracy of pairs found by \reremibkt and \reremi . Each dot represents a pair of columns, and its location indicates the highest-accuracy initial pair \reremibkt and \reremi .}
  \label{fig:scatter_reremi}
\end{figure}

\paragraph{Quality of the pairs.}
We compare the quality of the pairs found by \ownalgoinit and \reremibkt, the \reremi algorithm with pre-bucketing. \reremibkt gives more comparable running times and the quality of the initial pairs was essentially the same as with \reremi, as shown in Fig.~\ref{fig:scatter_reremi} for the \cms{$_{116395}$}, \cms{$_{7199}$} and \mamwc datasets. They show that \reremibkt and \reremi get the same results with the \mamwc dataset and there are no dots in the $x$-axis, which would indicate pairs found only by \reremi. With the \cms datasets, \reremi and \reremibkt find somewhat different initial pairs. \reremibkt finds almost always equally good initial pairs as \reremi, as well as many intial pairs that \reremi does not find. This indicates that comparison to \reremibkt for the quality of the initial pairs is fair.

\begin{figure}[tbp]
  \centering
  \includegraphics{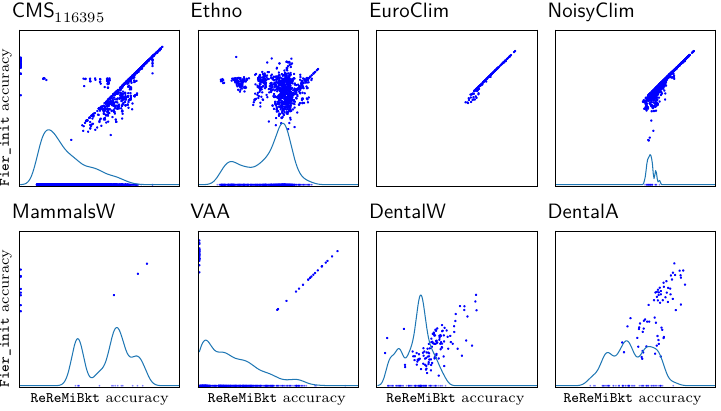}
  \caption{Comparing the accuracy of pairs found by \ownalgoinit and \reremibkt . Each dot represents a pair of columns, and its location indicates the highest-accuracy initial pair \ownalgoinit and \reremibkt found. The light blue line at the bottom shows the density of the dots that lie along the $x$-axis. All axes range across the unit interval.}
  \label{fig:scatter_all}
\end{figure}

Results on the \cms, \europeclimate, \noisyeurope, \ethno, \mamwc, \vaali, \dentalworld and \dentalasia datasets are shown in Fig.~\ref{fig:scatter_all}. 

The optimal outcome for these experiments is a diagonal line, indicating that \ownalgoinit finds pairs equivalent to those found by \reremibkt. This is mostly the case. With  \cms, \mamwc and \vaali, there are also some dots along the $x$-axis, meaning that there are pairs of columns that \ownalgo did not find, but most of those dots are at the left end of the $x$-axis, as indicated by the density plots (light blue). That is, they have low accuracy. \dentalworld and \dentalasia show weaker correlation, but still a clear diagonal pattern. \ethno is an outlier, where \ownalgoinit does not find some pairs of literals although they have reasonably high accuracy (indicated by the peak around the middle of the $x$-axis), but also finds pairs that have significantly higher accuracy than what \reremibkt returns. %

\paragraph{Speed of execution.}

The running times on different datasets are presented in Table~\ref{tab:runtimes}. The running times for \dentalworld and \dentalasia are not comparable to other datasets as they were carried out using a different setup (to reflect \cite{galbrun17computational,liu23emergence}). 

\begin{table}[tbp]
	\centering
	\caption{Comparing the running times (in seconds) of the algorithms on all datasets. 
        Running times with \dentalasia and \dentalworld are not comparable with others, as they were run using a different setup.}
        \label{tab:runtimes}
\begin{tabular}{
	@{}
	l@{\hspace{3ex}}
	*{7}{R}
	@{}
}
	\toprule
	Algorithm & \europeclimate & \noisyeurope & \ethno & \vaali & \dentalworld & \dentalasia \\ %
	\midrule
	\reremi & 1\,184.89 & 12\,922.03 & 9\,069.96 & 0.56 & 1\,569.01 & 1\,960.38 \\ %
	\reremibkt & 341.90 & 3\,163.01 & 102.70 &  0.68 & 320.09 & 233.02 \\ %
  \ownalgoinit & 10.50 & 68.52 & 13.65 & 1.87 & 20.66 & 18.24 \\ %
  \addlinespace
  \toprule
                 & \mamwc & \cms_{7199} & \cms_{14414} & \cms_{29136} & \cms_{58204} & \cms_{116395} \\
  \midrule
  \reremi &138.52 & 336.61 & 436.53 & 626.02 & 953.50 & 1\,961.12 \\
  \reremibkt &143.35 & 308.39 & 418.40 & 610.99 & 943.21 & 1\,899.83 \\
  \ownalgoinit &47.66 & 21.22 & 40.22 & 71.02 & 145.20 & 290.54 \\
	\bottomrule
\end{tabular}
\end{table}

Overall, we see that \ownalgoinit is consistently an order of magnitude faster than \reremibkt, and often two orders of magnitude faster than \reremi. The only exception to this is \vaali; the small size of the data and categorical attributes mean that the overhead of the hashing dominates the running time. Another case where the difference is smaller is \mamwc; here, the Boolean attributes mean less benefit for \ownalgoinit, and the on-the-fly discretisation of numerical attributes by \reremi is actually faster than pre-bucketing of \reremibkt. 

Looking at the different-sized \cms datasets, we also see that the running time of \ownalgoinit grows linearly with the number of rows, matching the asymptotic runtime. Even though \ownalgoinit uses more complex data structures for hashing, its memory consumption was only slightly larger than that of \reremi (\qty{\approx 30}{\percent}).

\paragraph{Sensitivity to parameters.}

We evaluated the sensitivity of \ownalgoinit to the parameters affecting locality-sensitive hashing. Overall, we found it to be very robust.

We tested all combinations of values $20$, $40$, $60$ and $80$ for $\bandsjacc$ and values $3$, $5$, $7$, $10$, $12$ and $15$ for $\rowsjacc$ and evaluated the results both w.r.t.\ running time and quality of answers. For these experiments we used the \europeclimate dataset. 

Table~\ref{tab:lsh_time} shows how the parameters impact the running time and accuracy. We can see that $\rowsjacc$ has the largest impact, the running time increasing with smaller values of $\rowsjacc$. This is because $\rowsjacc$ determines the length of the minhash signature, and shorter signatures imply a higher chance of matching and forming a candidate pair. On the other hand, the parameters do not have much impact on the average accuracy of the fifth-best result ($\bar{\jacc}@5$).

As the average quality does not change by much, we can conclude that the user can set rows and bands primarily based on how many initial pairs they want to find, but that the settings we have used throughout these experiments ($\rowsjacc = 10$, $\bandsjacc = 40$) seem to give a good balance. 

\begin{table}[tbp]
  \centering
  \caption{The effect of LSH parameters to running time (left) and average Jaccard of the fifth best result $\bar{\jacc}@5$ (right) depending on the number ($\bands$) and the width ($\rows$) of bands in LSH, for five repetitions and using \europeclimate data.}\label{tab:lsh_time}
  \begin{tabular}{
    @{}
    l
    @{\hspace*{2em}}
    *{6}{R}
    @{\hspace*{2em}}
    *{6}{R}
    @{}
    }
    \toprule
    & \multicolumn{6}{c}{time (s)} & \multicolumn{6}{c}{$\bar{\jacc}@5$} \\
    & \multicolumn{6}{c}{$\rowsjacc$} & \multicolumn{6}{c}{$\rowsjacc$}\\
    \cmidrule(r{2em}){2-7} \cmidrule{8-13}
    $\bandsjacc$ & {3} & {5} & {7} & {10} & {12} & {15} & {3} & {5} & {7} & {10} & {12} & {15}\\
    \midrule
    20 & 33.89 & 21.34 & 15.00 & 6.94 & 5.68 & 5.14
         &0.69 & 0.70 & 0.63 & 0.69 & 0.50 & 0.63 \\
    40 & 45.71 & 33.22 & 23.09 & 13.46 & 11.30 & 9.54
         & 0.71 & 0.73 & 0.65 & 0.72 & 0.57 & 0.67 \\
    60 & 56.91 & 40.77 & 29.79& 18.59 & 15.66 & 13.88
         & 0.73 & 0.73 & 0.65 & 0.71 & 0.62 & 0.70 \\
    80 & 68.86 & 48.69 & 35.61 & 23.76 & 20.35 & 17.73
         & 0.72 & 0.71 & 0.66 & 0.73 & 0.63 & 0.70 \\
	\bottomrule
\end{tabular}
\end{table}

\subsection{Extending Initial Pairs}
\label{sec:extend-init-pairs}

The next phase of the \ownalgo algorithm is the extension of the initial pairs. To test this phase, we use the same pre-mined initial pairs for all algorithms and compare extensions obtained with \ownalgoext versus with the exhaustive attempts of \reremi. For the extensions, pre-bucketing as done by \reremibkt is not expected to bring any benefits for \reremi, as its cut-point algorithm is very efficient (cf.\ Table~\ref{tab:runtimes}). \ownalgoext might extend some initial pairs with literals that are somewhat inferior to the ones found by \reremi, or it might fail to find any extensions at all. On the other hand, it should be much faster than \reremi.
For these tests we only consider \mamwc, \dentalworld and two \cms datasets due to space reasons. 

\begin{figure}[tpb]
  \centering
  \includegraphics{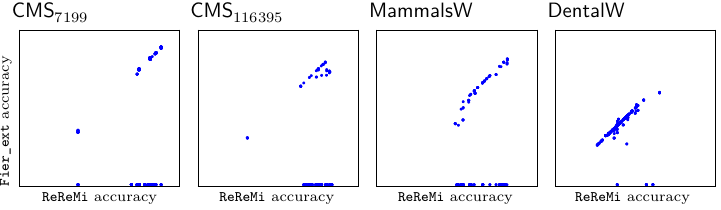}
  \caption{Comparing the accuracy of once extended initial pairs by \ownalgoext and \reremi . Each dot represents a pair of columns, and its location indicates the highest-accuracy initial pair \ownalgo and \reremi found. All axes range across the unit interval.}
  \label{fig:scatter_one_ext}
\end{figure}

\paragraph{Extending once.}
In the first test, we check how similar the extensions are after one round of extensions. The results are presented in Fig.~\ref{fig:scatter_one_ext}. This plot is similar to Fig.~\ref{fig:scatter_all}. As we can see, there are some initial pairs for which \ownalgoext does not find any valid extension, but for those that it does find, the results are generally as good as those of \reremi. %

\begin{table}[p]
  \centering
  \caption{Accuracy, total number of extension steps, and time in seconds when extending the \reremi initial pairs multiple times. All results are averages over 5 runs.}\label{tab:reremi_init_extension}
  \begin{tabular}{
    @{}
    l%
    *{8}{R}
    @{}
}
    \toprule
    & & & \multicolumn{3}{c}{\mamwc} & \multicolumn{3}{c}{\dentalworld}\\
    \cmidrule(r){4-6} \cmidrule(l){7-9}
    algorithm & \multicolumn{1}{c}{$\rowsham$} & \multicolumn{1}{c}{$\bandsham$} & \multicolumn{1}{c}{$\jacc@10$} & \multicolumn{1}{c}{\# ext.} & \multicolumn{1}{c}{time (s)} &  \multicolumn{1}{c}{$\jacc@10$} & \multicolumn{1}{c}{\# ext.} & \multicolumn{1}{c}{time (s)} \\
    \midrule
    \ownalgoext & 10 &  10 & 0.77 & 296.4 & 1471.74 & 0.60 & 535.8 & 643.16\\
			& & 20 & 0.79 & 329.0 & 1664.60 & 0.61 & 547.8 & 712.50\\
			& & 40 & 0.79 & 354.8 & 1764.90 & 0.62 & 543.8 & 776.40\\
			& 20 &  10 & 0.72 & 112.6 & 768.05 & 0.57 & 500.2 & 390.47\\
			& & 20 & 0.72 & 130.6 & 973.91 & 0.58 & 523.0 & 472.29\\
			& & 40 & 0.75 & 237.4 & 1383.91 & 0.59 & 530.6 & 578.33\\
			& 30 &  10 & 0.68 & 60.0 & 438.41 & 0.54 & 402.8 & 245.93\\
			& & 20 & 0.70 & 65.4 & 641.48 & 0.57 & 462.2 & 348.41\\
			& & 40 & 0.71 & 101.0 & 990.65 & 0.57 & 512.8 & 483.04\\
			& 40 &  10 & 0.63 & 38.4 & 285.89 & 0.51 & 229.0 & 125.20\\
			& & 20 & 0.68 & 54.6 & 473.24 & 0.53 & 329.0 & 221.54\\
			& & 40 & 0.70 & 62.0 & 727.29 & 0.54 & 426.2 & 395.37\\
    [.5em]
    \reremi  & & & 0.83 & 410.0 & 3263.28 & 0.61 & 517.0 & 1494.44\\
    \addlinespace
    \toprule
    & & & \multicolumn{3}{c}{\cms{$_{7199}$}} & \multicolumn{3}{c}{\cms{$_{116395}$}}\\
    \cmidrule(r){4-6} \cmidrule(l){7-9}
    algorithm & \multicolumn{1}{c}{$\rowsham$} & \multicolumn{1}{c}{$\bandsham$} & \multicolumn{1}{c}{$\jacc@10$} & \multicolumn{1}{c}{\# ext.} & \multicolumn{1}{c}{time (s)} &  \multicolumn{1}{c}{$\jacc@10$} & \multicolumn{1}{c}{\# ext.} & \multicolumn{1}{c}{time (s)} \\
    \midrule
    \ownalgoext& 10 &  10 & 0.82 & 28.8 & 90.08 & 0.77 & 90.0 & 1072.10\\
& & 20 & 0.82 & 40.0 & 181.25 & 0.78 & 106.8 & 1906.59\\
& & 40 & 0.82 & 47.0 & 299.37 & 0.80 & 126.4 & 2646.13\\
& 20 &  10 & 0.80 & 4.0 & 7.56 & 0.72 & 11.4 & 202.65\\
& & 20 & 0.81 & 6.4 & 17.68 & 0.73 & 20.4 & 303.56\\
& & 40 & 0.81 & 8.6 & 31.13 & 0.74 & 41.8 & 532.56\\
& 30 &  10 & 0.80 & 0.6 & 6.82 & 0.71 & 5.0 & 148.69\\
& & 20 & 0.80 & 1.4 & 12.58 & 0.72 & 5.6 & 228.74\\
& & 40 & 0.81 & 3.0 & 23.88 & 0.72 & 8.4 & 421.66\\
& 40 &  10 & 0.79 & 0.0 & 7.35 & 0.71 & 1.4 & 117.36\\
& & 20 & 0.79 & 0.6 & 13.86 & 0.71 & 2.8 & 223.92\\
& & 40 & 0.80 & 0.8 & 25.98 & 0.71 & 2.8 & 396.22\\[.5em]

    \reremi  & & & 0.83 & 55.0 & 695.85 & 0.81 & 154.0 & 7626.95\\

    \bottomrule
  \end{tabular}
\end{table}
\begin{table}[tb]
  \centering
  \caption{Accuracy, number of extension steps, number of resulting redescriptions and time in seconds for full redescription mining. All results are averages over 5 runs.}\label{tab:lsh_init_extension}
  \begin{tabular}{
    @{}
    l%
    *{10}{R}
    @{}
}
    \toprule
    & & & \multicolumn{4}{c}{\mamwc} & \multicolumn{4}{c}{\dentalworld}\\
    \cmidrule(r){4-7} \cmidrule(l){8-11}
    alg. & \multicolumn{1}{c}{$\rowsham$} & \multicolumn{1}{c}{$\bandsham$} &  \multicolumn{1}{c}{$\jacc@10$} & \multicolumn{1}{c}{\# ext.}  &  \multicolumn{1}{c}{\# results} & \multicolumn{1}{c}{time} &  \multicolumn{1}{c}{$\jacc@10$} & \multicolumn{1}{c}{\# ext.}  & \multicolumn{1}{c} {\# results} & \multicolumn{1}{c}{time} \\
    \midrule
    \ownalgofull & 20 & 20 & 0.74 & 94.6 & 32.4& 765 & 0.57 & 486.8 & 216.8 & 445\\
& & 40 & 0.72 & 75.0 & 24.0& 901 & 0.59 & 505.8 & 222.6 & 549\\
& 30 & 20 & 0.63 & 39.0 & 14.2& 446 & 0.52 & 423.8 & 189.0 & 315\\
& & 40 & 0.69 & 66.8 & 20.8& 705 & 0.55 & 454.6 & 202.6 & 447\\
& 40 & 20 & 0.63 & 28.0 & 10.6& 333 & 0.51 & 258.8 & 139.6 & 199\\
& & 40 & 0.58 & 37.0 & 13.2& 500 & 0.53 & 362.8 & 178.2 & 358\\[.5em]
    \reremi  & & & 0.83 & 410.0 & 71.0 & 3401 & 0.61 & 517.0 & 93.0 & 3063\\
    \reremibkt & & & 0.83& 410.0	&71.0 & 3381 & 0.63& 512.0&90.0&1866\\
    \addlinespace
    \toprule
    & & & \multicolumn{4}{c}{\cms{$_{7199}$}} & \multicolumn{4}{c}{\cms{$_{116395}$}}\\
    \cmidrule(r){4-7} \cmidrule(l){8-11}
    alg. & \multicolumn{1}{c}{$\rowsham$} & \multicolumn{1}{c}{$\bandsham$} &  \multicolumn{1}{c}{$\jacc@10$} & \multicolumn{1}{c}{\# ext.}  &  \multicolumn{1}{c}{\# results} & \multicolumn{1}{c}{time} &  \multicolumn{1}{c}{$\jacc@10$} & \multicolumn{1}{c}{\# ext.}  & \multicolumn{1}{c} {\# results} & \multicolumn{1}{c}{time} \\
    \midrule
    \ownalgofull& 20 & 20 & 0.75 & 12.2 & 18.8& 34 & 0.76 & 19.6 & 21.4 & 463\\
& & 40 & 0.77 & 23.2 & 23.0& 53 & 0.76 & 22.8 & 23.8 & 592\\
& 30 & 20 & 0.74 & 3.6 & 17.0& 30 & 0.75 & 4.2 & 17.2 & 381\\
& & 40 & 0.75 & 8.8 & 19.8& 41 & 0.75 & 6.4 & 18.4 & 517\\
& 40 & 20 & 0.73 & 3.0 & 17.6& 31 & 0.73 & 0.6 & 16.0 & 367\\
& & 40 & 0.72 & 2.4 & 17.2& 41 & 0.74 & 1.4 & 16.4 & 510\\[.5em]

    \reremi  & & & 0.83 & 55.0 & 22.0 & 1032 & 0.81 & 154.0 & 45.0 & 96\\
    \reremibkt & & &0.83	&55.0	&22.0	&957&0.83&79.0&	29.0	&5796\\
    \bottomrule
  \end{tabular}
\end{table}
 
\paragraph{Extending multiple times.}
Genuine redescription mining involves extending the pairs more than just once. At this point it is no longer sensible to compare the accuracies of individual initial pairs, as small differences in each extension step can yield vastly different redescriptions, and the greedy extension done by \reremi can sometimes yield worse results than \ownalgoext. Instead, we consider the average Jaccard of the 10th best result ($\bar{\jacc}@10$). We also measure the number of extension steps done by the algorithms \emph{in total}. That is, if there are $100$ initial pairs, and each pair is extended $3$ times on both sides, the total number of extension steps is $600$. Notice that \reremi stops extensions when it cannot find any literal to extend with. The results of this experiment are presented in Table~\ref{tab:reremi_init_extension}. Further parameter combinations are presented in Appendix~\ref{sec:app-extend-init-pairs}.  %

As can be seen in Table~\ref{tab:reremi_init_extension}, the average $\bar{\jacc}@10$ for \ownalgoext is somewhat less than for \reremi. With appropriate parameters, however, the difference is small. On the other hand, the running times can differ significantly.  With \dentalworld for instance, \reremi takes almost \num{1500} seconds while \ownalgoext achieves the same accuracy in half the time (using $\rowsham=10, \bandsham=20$), or comparable accuracy in less than a quarter of the time (using $\rowsham=30, \bandsham=20$). We can also see that for the extensions, smaller values of $\rowsham$ yield higher running times. This is because with smaller $\rowsham$, LSH is more random and tends to generate much more potential extensions. On the other hand, higher values of $\bandsham$ tend to increase the accuracy, as LSH then simply does more repetitions.

\subsection{Building Full Redescriptions}
\label{sec:exp-full}

The final test is for the full algorithm \ownalgofull, where we use LSH for both initial pairs and extensions. The results are presented in Table~\ref{tab:lsh_init_extension} with further results in Appendix~\ref{sec:app-exp-full}. %

For the \cms datasets, \ownalgofull cannot find many extensions. This situation is the same as in Table~\ref{tab:reremi_init_extension}, indicating that it is probably a feature of the extension algorithm rather than of the initial pairs. On the other hand, the average accuracies are still quite high, indicating that the lack of extensions might be a consequence of having high-accuracy initial pairs that cannot be extended with higher accuracy. The running times of the algorithm are very low, showing a significant improvement over \reremi and \reremibkt.

\section{Conclusions}
\label{sec:conclusions}

Locality-sensitive hashing in \ownalgo significantly speeds up finding the initial pairs and extending them, without sacrificing the quality. Handling intervals from numerical variables requires a more complex approach, especially for the initial pairs, but it pays off with notably improved running times. Based on our experiments, we can recommend using \ownalgo for the initialisation of greedy redescription mining without reservation. The extension phase is somewhat more sensitive to hyperparameters and is usually not as significant a bottleneck. For larger data sets or for quick or interactive testing, we can recommend \ownalgofull.

\bibliographystyle{splncs04}
\newcommand\textdot{\.}

\appendix

\section{Further Experimental Evaluation}
\label{sec:app-exper-eval}

\subsection{Extending Initial Pairs}
\label{sec:app-extend-init-pairs}

\paragraph{Extending multiple times.}
The full results for extending pre-mined initial pairs are presented in Table~\ref{tab:app_reremi_init_extension}.

\begin{table}[tbp]
  \centering
  \caption{Accuracy, total number of extensions, and time in seconds when extending the \reremi initial pairs multiple times. All results are averages over 5 runs.}\label{tab:app_reremi_init_extension}
  \begin{tabular}{
    @{}
    l%
    *{2}{S[table-format=2.0]}%
    S[table-format=1.2,table-auto-round,separate-uncertainty]%
    S[table-format=3.1,table-auto-round,separate-uncertainty]%
    S[table-format=3.2,table-auto-round,separate-uncertainty]%
    S[table-format=1.2,table-auto-round,separate-uncertainty]%
    S[table-format=3.1,table-auto-round,separate-uncertainty]%
    S[table-format=3.2,table-auto-round,separate-uncertainty]%
    @{}
}
    \toprule
    & & & \multicolumn{3}{c}{\mamwc} & \multicolumn{3}{c}{\dentalworld}\\
    \cmidrule(r){4-6} \cmidrule(l){7-9}
    alg. & {$\rowsham$} & {$\bandsham$} & {$\jacc@10$} & {\# of ext.} & {time (s)} &  {$\jacc@10$} & {\# of ext.} & {time (s)} \\
    \midrule
    \ownalgoext & 10 & 10 & 0.7714 & 296.4 & 1471.736 & 0.6042 & 535.8 & 643.1600000000001\\
    & & 20 & 0.788 & 329.0 & 1664.6 & 0.609 & 547.8 & 712.504\\
    & & 40 & 0.7862 & 354.8 & 1764.898 & 0.6152 & 543.8 & 776.4\\
    & & 80 &  0.8244 & 447.8 & 2176.3419999999996 & 0.6135999999999999 & 545.6 & 819.516\\
    & 20 & 10 & 0.718 & 112.6 & 768.05 & 0.5725999999999999 & 500.2 & 390.47\\
    & & 20 & 0.7213999999999999 & 130.6 & 973.9100000000001 & 0.5841999999999999 & 523.0 & 472.288\\
    & & 40 & 0.7468 & 237.4 & 1383.914 & 0.5919999999999999 & 530.6 & 578.3340000000001\\
    & & 80 & 0.7522 & 259.6 & 1674.146 & 0.5986 & 541.0 & 727.704\\
    & 30 & 10 & 0.6778 & 60.0 & 438.40600000000006 & 0.5424 & 402.8 & 245.92600000000002\\
    & & 20 & 0.7003999999999999 & 65.4 & 641.48 & 0.5678 & 462.2 & 348.40999999999997\\
    & & 40 & 0.7132 & 101.0 & 990.6479999999999 & 0.5723999999999999 & 512.8 & 483.03599999999994\\
    & & 80 & 0.7212 & 133.8 & 1369.4240000000002 & 0.5772 & 519.0 & 686.5160000000001\\
    & 40 & 10 & 0.6268 & 38.4 & 285.886 & 0.5128 & 229.0 & 125.196\\
    & & 20 & 0.6836 & 54.6 & 473.238 & 0.5344 & 329.0 & 221.538\\
    & & 40 & 0.697 & 62.0 & 727.286 & 0.538 & 426.2 & 395.374\\
    & & 80 & 0.7011999999999999 & 63.6 & 1162.734 & 0.5658 & 480.0 & 658.6719999999999\\
    & 50 & 10 & 0.6202 & 34.6 & 209.512 & 0.48179999999999995 & 105.4 & 79.45200000000001\\
    & & 20 & 0.627 & 40.8 & 333.932 & 0.49939999999999996 & 168.4 & 133.85999999999999\\
    & & 40 & 0.6659999999999999 & 53.8 & 677.988 & 0.5164 & 255.0 & 256.28200000000004\\
    & & 80 & 0.6926 & 57.2 & 1122.534 & 0.538 & 349.8 & 538.832\\
    \reremi  & & & 0.829 & 410.0 & 3263.28 & 0.609 & 517.0 & 1494.44\\
    \addlinespace
    \toprule
    & & & \multicolumn{3}{c}{\cms{$_{7199}$}} & \multicolumn{3}{c}{\cms{$_{116395}$}}\\
    \cmidrule(r){4-6} \cmidrule(l){7-9}
    alg. & {$\rowsham$} & {$\bandsham$} & {$\jacc@10$} & {\# of ext.} & {time (s)} &  {$\jacc@10$} & {\# of ext.} & {time (s)} \\
    \midrule
    \ownalgoext & 10 & 10 & 0.8177999999999999 & 28.8 & 90.08400000000002 & 0.77 & 90.0 & 1072.096\\
    & & 20 & 0.8188000000000001 & 40.0 & 181.25 & 0.7758 & 106.8 & 1906.5880000000002\\
    & & 40 & 0.8221999999999999 & 47.0 & 299.37 & 0.7984 & 126.4 & 2646.1279999999997\\
    & & 80 & 0.8228 & 52.4 & 383.05 & 0.8030000000000002 & 127.6 & 4366.603999999999\\
    & 20 & 10 & 0.8038000000000001 & 4.0 & 7.564 & 0.7194 & 11.4 & 202.652\\
    & & 20 & 0.8058 & 6.4 & 17.676 & 0.7303999999999999 & 20.4 & 303.56399999999996\\
    & & 40 & 0.8139999999999998 & 8.6 & 31.131999999999998 & 0.7402 & 41.8 & 532.5600000000001\\
    & & 80 & 0.8164 & 14.6 & 60.589999999999996 & 0.752 & 55.0 & 964.702\\
    & 30  & 10 & 0.7958000000000001 & 0.6 & 6.82 & 0.7101999999999999 & 5.0 & 148.69\\
    & & 20 & 0.8009999999999999 & 1.4 & 12.575999999999999 & 0.7177999999999999 & 5.6 & 228.744\\
    & & 40 & 0.8098000000000001 & 3.0 & 23.876 & 0.7197999999999999 & 8.4 & 421.65600000000006\\
    & & 80 & 0.8116 & 4.6 & 46.206 & 0.7236 & 9.4 & 730.606\\
    & 40 & 10 & 0.791 & 0.0 & 7.354000000000001 & 0.706 & 1.4 & 117.35999999999999\\
    & & 20 & 0.7942 & 0.6 & 13.857999999999999 & 0.7085999999999999 & 2.8 & 223.916\\
    & & 40 & 0.7982 & 0.8 & 25.977999999999998 & 0.706 & 2.8 & 396.21999999999997\\
    & & 80 & 0.7964 & 0.8 & 49.28600000000001 & 0.7133999999999999 & 4.6 & 732.576\\
    & 50 & 10 & 0.791 & 0.0 & 7.975999999999999 & 0.706 & 0.6 & 120.48600000000002\\
    & & 20 & 0.791 & 0.0 & 14.9 & 0.706 & 0.8 & 197.45\\
    & & 40 & 0.7942 & 0.4 & 28.248 & 0.7061999999999999 & 1.6 & 368.97799999999995\\
    & & 80 & 0.797 & 0.8 & 55.20399999999999 & 0.7085999999999999 & 2.8 & 730.6720000000001\\
    \reremi  & & & 0.827 & 55.0 & 695.85 & 0.807 & 154.0 & 7626.95\\
    \bottomrule
  \end{tabular}
\end{table}

\subsection{Building Full Redescriptions}
\label{sec:app-exp-full}

The results are presented in Table~\ref{tab:app_lsh_init_extension}.

\begin{table}[tbp]
  \centering
  \caption{Accuracy, total number of extensions, total number of resulting redescriptions, and total time in seconds for full redescription mining. All results are averages over 5 runs.}\label{tab:app_lsh_init_extension}
  \begin{tabular}{
    @{}
    l%
    *{2}{S[table-format=2.0]}%
    S[table-format=1.2,table-auto-round,separate-uncertainty]%
    S[table-format=2.1,table-auto-round,separate-uncertainty]%
    S[table-format=2.1,table-auto-round,separate-uncertainty]%
    S[table-format=4.0,table-auto-round,separate-uncertainty]%
    S[table-format=1.2,table-auto-round,separate-uncertainty]%
    S[table-format=3.1,table-auto-round,separate-uncertainty]%
    S[table-format=3.1,table-auto-round,separate-uncertainty]%
    S[table-format=4.0,table-auto-round,separate-uncertainty]%
    @{}
}
    \toprule
    & & & \multicolumn{4}{c}{\mamwc} & \multicolumn{4}{c}{\dentalworld}\\
    \cmidrule(r){4-7} \cmidrule(l){8-11}
    alg. & {$\rowsham$} & {$\bandsham$} & {$\jacc@10$} & {\# ext.} & {\# results} & {time} &  {$\jacc@10$} & {\# ext.}& {\# results} & {time} \\
    \midrule
    \ownalgofull & 20 & 10 & 0.6794 & 60.4 & 19.2 & 517.9100000000001 & 0.5514 & 446.8& 194.6 & 345.454\\
    & & 20 & 0.742 & 94.6 & 32.4 & 764.646 & 0.5702 & 486.8& 216.8 & 444.75200000000007\\
    & & 40 & 0.7203999999999999 & 75.0 & 24.0 & 900.902 & 0.5854 & 505.8& 222.6 & 548.526\\
    & 30 & 10 & 0.6022000000000001 & 33.2 & 12.4 & 285.202 & 0.5322 & 348.2& 169.0 & 216.94400000000002\\
    & & 20 & 0.6316 & 39.0 & 14.2 & 445.63199999999995 & 0.524 & 423.8& 189.0 & 315.474\\
    & & 40 & 0.6888000000000001 & 66.8 & 20.8 & 704.894 & 0.5514 & 454.6& 202.6 & 447.442\\
    & 40 & 10 & 0.5572 & 28.2 & 13.0 & 213.42800000000003 & 0.47759999999999997 & 227.6& 124.0 & 144.46599999999998\\
    & & 20 & 0.6332 & 28.0 & 10.6 & 332.844 & 0.5094000000000001 & 258.8& 139.6 & 199.18\\
    & & 40 & 0.5768000000000001 & 37.0 & 13.2 & 500.09 & 0.5274000000000001 & 362.8& 178.2 & 357.586\\
    \reremi  & & & 0.827 & 410 & 71 & 3401.796 & 0.609 & 517 & 93 & 3063.450799\\
    \reremibkt & & & 0.829& 410	&71 & 3381.1800000000003 & 0.63& 512&90&1866.48\\
    \addlinespace
    \toprule
    & & & \multicolumn{4}{c}{\cms{$_{7199}$}} & \multicolumn{4}{c}{\cms{$_{116395}$}}\\
    \cmidrule(r){4-7} \cmidrule(l){8-11}
    alg. & {$\rowsham$} & {$\bandsham$} & {$\jacc@10$} & {\# ext.} & {\# results} & {time} &  {$\jacc@10$} & {\# of ext.}& {\# results} & {time} \\
    \midrule
    \ownalgofull & 20 & 10 & 0.7442 & 8.8 & 18.4 & 30.137999999999998 & 0.7416 &9.6 & 19.2 & 352.42\\
    & & 20 & 0.7502000000000001 & 12.2 & 18.8 & 33.986000000000004 & 0.7558 &19.6 & 21.4 & 462.912\\
    & & 40 & 0.7654 & 23.2 & 23.0 & 52.64 & 0.7614000000000001 &22.8 & 23.8 & 591.6659999999999\\
    & 30 & 10 & 0.7416 & 2.0 & 19.8 & 23.75 & 0.7253999999999999 &1.6 & 17.6 & 302.238\\
    & & 20 & 0.7426 & 3.6 & 17.0 & 30.312 & 0.7458 &4.2 & 17.2 & 380.928\\
    & & 40 & 0.7498 & 8.8 & 19.8 & 41.31 & 0.7486 &6.4 & 18.4 & 517.136\\
    & 40 & 10 & 0.7332000000000001 & 0.6 & 15.0 & 24.004 & 0.7236 &1.0 & 15.4 & 295.44\\
    & & 20 & 0.7312 & 3.0 & 17.6 & 30.812 & 0.7338 &0.6 & 16.0 & 366.61400000000003\\
    & & 40 & 0.7245999999999999 & 2.4 & 17.2 & 40.564 & 0.7358 &1.4 & 16.4 & 510.04200000000003\\
    \reremi  & & & 0.827 & 55 & 22 & 1032.46 & 0.807 & 154 & 45 & 9588.07\\
    \reremibkt & & &0.827	&55	&22	&957.24&0.83&79&	29	&5796.780000000001\\
    \addlinespace
    \toprule
    & & & \multicolumn{4}{c}{\ethno} & \multicolumn{4}{c}{\vaali}\\
    \cmidrule(r){4-7} \cmidrule(l){8-11}
    alg. & {$\rowsham$} & {$\bandsham$} & {$\jacc@10$} & {\# ext.} & {\# results} & {time} &  {$\jacc@10$} & {\# of ext.}& {\# results} & {time} \\
    \midrule
    \ownalgofull & 20 & 10 & 0.5542 & 79.2 & 50.2 & 38.266 & 0 & 0& 0 & 0\\
    & & 20 & 0.5529999999999999 & 105.6 & 55.8 & 53.456 & 0.389 & 0.0& 1.0 & 1.57\\
    & & 40 & 0.5538000000000001 & 137.4 & 66.2 & 73.63000000000001 & 0 & 0& 0 & 0\\
    & 30 & 10 & 0.49420000000000003 & 30.0 & 37.0 & 14.85 & 0 & 0& 0 & 0\\
    & & 20 & 0.5094000000000001 & 39.4 & 34.8 & 21.962 & 0 & 0& 0 & 0\\
    & & 40 & 0.5464 & 69.6 & 45.4 & 33.822 & 0 & 0& 0 & 0\\
    & 40 & 10 & 0.4784 & 6.0 & 27.8 & 9.84 & 0 & 0& 0 & 0\\
    & & 20 & 0.454 & 13.0 & 25.0 & 12.224 & 0.356 & 0.0& 1.0 & 1.71\\
    & & 40 & 0.4988 & 26.6 & 35.2 & 17.568 & 0 & 0& 0 & 0\\
    \reremi  & & & 0.652 & 13 & 3 & 184.417 & 0.39 & 6 & 4 & 7.68\\
    \bottomrule
  \end{tabular}
\end{table}

\end{document}